\documentclass[pdflatex,iicol,sn-basic]{sn-jnl}

\usepackage{arydshln}
\DeclareMathSymbol{\unaryminus}{\mathord}{AMSa}{"39}
\newsavebox\CBox

\jyear{2022}%

\begin{document}

\title[~]{On Fitness Landscape Analysis of Permutation Problems: From Distance Metrics to Mutation Operator Selection}


\author*[1]{\fnm{Vincent A.} \sur{Cicirello}}\email{cicirelv@stockton.edu}

\affil*[1]{\orgdiv{Computer Science}, \orgname{Stockton University}, \orgaddress{\street{101 Vera King Farris Dr}, \city{Galloway}, \state{NJ}, \postcode{08205}, \country{USA}}}

\abstract{In this paper, we explore the theory and expand upon the practice of fitness
landscape analysis for optimization problems over the space of permutations. Many of the 
computational and analytical tools for fitness landscape analysis, such as fitness distance
correlation, require identifying a distance metric for measuring the similarity of different 
solutions to the problem. We begin with a survey of the available distance metrics for 
permutations, and then use principal component analysis to classify these metrics. The result 
of this analysis aligns with existing classifications of permutation problem types produced 
through less formal means, including the A-permutation, R-permutation, and P-permutation types, 
which classifies problems by whether absolute position of permutation elements, relative positions 
of elements, or general precedence of pairs of elements, is the dominant influence 
over solution fitness. Additionally, the formal analysis identifies subtypes within these problem 
categories. We see that the classification can assist in identifying appropriate metrics based on 
optimization problem feature for use in fitness landscape analysis. Using optimization problems 
of each class, we also demonstrate how the classification scheme can subsequently inform the 
choice of mutation operator within an evolutionary algorithm. From this, we present a classification
of a variety of mutation operators as a counterpart to that of the metrics. Our implementations 
of the permutation metrics, permutation mutation operators, and associated evolutionary algorithm, are 
available in a pair of open source Java libraries. All of the code necessary to recreate our 
analysis and experimental results are also available as open source.}



\maketitle

\section{Introduction}\label{sec:intro}

Analyses of evolutionary algorithms (EA) and other forms of optimization often employ
a fitness landscape~\citep{mitchell}, which is a spatial arrangement of the space of all possible solutions to the
problem such that structurally ``similar'' solutions are located near to each other. To be most relevant
to the algorithm under analysis, nearby solutions should correspond with solutions reachable in a small
number of applications of the search operators. Thus, an effective fitness landscape analysis requires
identifying an appropriate measure of distance for the combination of problem and operators. There exists
much work on fitness landscape analysis, including for permutation 
landscapes~\citep{hernando2015,tayarani2014,cicirello2014,cicirello2013,sorensen07,schiavinotto2007,Reidys2002}.
This current article is an extended version of our prior conference paper on the topic~\citep{cicirello2019}.

The available fitness landscape analysis tools include
fitness distance correlation (FDC)~\citep{fdc}, Pearson correlation between 
solution fitness and distance to the nearest optimal solution, as well as 
search landscape calculus~\citep{cicirello2016}, which examines the local rate of change of fitness.
FDC, search landscape calculus, and other related techniques require
distance metrics. The features of a permutation, or other structure, that are 
important in determining similarity or distance is often problem dependent.  
For the Traveling Salesperson Problem (TSP), the permutation represents a set of 
edges between adjacent pairs of cities, and is
thus rotationally invariant since rotation doesn't change element
adjacency. Permutations can also represent one-to-one mappings between sets,
such as in the largest common subgraph problem, where one seeks the largest
subgraph of graph $G_1$ that is isomorphic to a subgraph of $G_2$. You can 
represent a mapping by ordering the vertexes of $G_1$,
and using a permutation of the vertexes of $G_2$, such that
vertex $i$ of $G_2$ in permutation index $j$ corresponds to mapping 
vertex $i$ of $G_2$ to vertex $j$ of $G_1$. In this case, absolute element position 
is most important to fitness. \cite{campos2005} categorized 
permutation problems into two types: R-permutation, 
such as the TSP, where relative positions (i.e., adjacency implies edges) are important; 
and A-permutation, such as mapping problems, where absolute positions have 
greatest effect on fitness. \cite{cicirello2016} previously added a third type, P-permutation, 
where general element precedences most directly impact 
fitness (e.g., element $w$ occurs prior to elements $x$, $y$, and $z$, but not 
necessarily adjacent to any of them). Many scheduling problems are this type 
(e.g., a job may be delayed if jobs with long process times are anywhere prior 
to it in the schedule).

We survey permutation distances in Section~\ref{sec:distance}, and then formally 
identify groups of related permutation metrics using principal component analysis (PCA) in 
Sections~\ref{sec:classes} and~\ref{sec:longperms}. The first three principal components 
correspond to the three problem classes defined above; and the next few identify new subtypes.  
A classification of metrics aligned with the existing classification of  
problems is a desirable property. For example, if one requires a metric relevant for analyzing 
the fitness landscape of a problem within a known problem class, then the distance 
classification can directly lead to the most relevant metrics. In Section~\ref{sec:landscapes}, we 
provide a set of fitness landscapes corresponding to the identified classes of permutation metric. For each
landscape and metric, we compute FDC as an example application of the classification scheme.
We implement the PCA, as well as our FDC examples, using JavaPermutationTools, an open source library of permutation 
distance metrics~\citep{cicirello2018}. In Section~\ref{sec:mutation}, we apply the
classification to choosing a mutation operator for use in an EA, and derive a classification of 
a wide variety of permutation mutation operators. The experiments use the EA
implementation of the open source Chips-n-Salsa library~\citep{cicirello2020}. The code to replicate the
experiments and analysis is also available: \url{https://github.com/cicirello/MONE2022-experiments}.
We wrap up with a discussion in Section~\ref{sec:conclude}.

\section{Permutation Distance}\label{sec:distance}

Table~\ref{tab:summary} summarizes the permutation distances used in the analysis,
including runtime, and metric status. The $n$ in runtimes and 
equations is permutation length, $p(i)$ refers to
the element at index $i$ in permutation $p$, $p^{\unaryminus\!1}(e)$ refers to the 
index of element $e$ in $p$. We use 1-based indexing in equations
(index of first position is 1). Subscripts refer to different permutations.  
Thus, $p_1(i)$ refers to the element in position $i$ of permutation $p_1$.

\begin{table}[t]
\begin{center}
\begin{minipage}{216pt} %%{174pt}
\caption{Summary of distance measure classes}\label{tab:summary}
\begin{tabular}{@{}lll@{}}
\toprule
Distance		& Runtime	& Metric? \\ 
\midrule
edit	& $O(n^2)$	& yes \\
exact match	& $O(n)$	& yes \\
interchange	& $O(n)$	& yes \\
acyclic edge	& $O(n)$	& pseudo \\
cyclic edge	& $O(n)$	& pseudo \\
r-type 		& $O(n)$	& yes \\
cyclic r-type	& $O(n)$	& pseudo \\
Kendall tau	& $O(n \lg n)$	& yes \\
reinsertion	& $O(n \lg n)$	& yes \\
deviation (dev)	& $O(n)$	& yes \\
normalized dev. & $O(n)$	& yes \\
squared dev. & $O(n)$	& yes \\
Lee		& $O(n)$	& yes \\
reversal edit & init: $O(n! n^3)$ & yes \\
   & 	calc: $O(n^2)$ & \\
\botrule
\end{tabular}
\end{minipage}
\end{center}
\end{table}

\textbf{Edit distance:}
The edit distance is the minimum cost of the ``edit operations'' 
to transform one structure into another.
Levenshtein distance is a string edit distance~\citep{levenshtein1966}, 
on binary strings (i.e., of ones and zeros),
with edit operations character insertion, removal, and 
changes. \cite{wagner74} extended 
this to non-binary strings, introduced  
configurable costs to the three edit operations, and provided a dynamic 
programming algorithm to compute it.
\cite{sorensen07} suggested treating a permutation as a string, 
and applying string edit distance. All edit distances are metrics. Our  
implementation is of Wagner and Fischer's dynamic programming algorithm,
including parameters for the costs of the edit operations.  
Runtime is $O(n^2)$.
   
\textbf{Exact match distance:}
\cite{ronald1998} extends Hamming distance to non-binary strings, producing
exact match distance, an edit distance with element changes as the edit operation.
It is widely used~\citep{cicirello2014,cicirello2013,sorensen07,sevaux2005},
satisfies the metric properties~\citep{ronald1998}, and computed in $O(n)$ time
by counting the number of positions with different elements:
\begin{equation*}
\delta(p_1, p_2) = \lvert \{ i\in\{1 \ldots n\} \mid p_1(i) \neq p_2(i) \} \rvert .
%\sum_{i\in\{1 \ldots n\}, \; p_1(i) \neq p_2(i)} 1 . 
\end{equation*}

\textbf{Interchange distance:}
Interchange distance is an edit distance with one edit operation, element interchanges.
It is the minimum number of swaps needed to transform $p_1$ into $p_2$; and
is computed efficiently ($O(n)$ time) by counting the number of cycles 
between the permutations~\citep{cicirello2013}.
A permutation cycle of length $k$ is transformed into $k$ fixed points with $k-1$ swaps 
(a fixed point is a cycle of length 1). Let $\textnormal{CycleCount}(p_1,p_2)$ be the number of permutation cycles,
and define interchange distance as:
\begin{equation*}\label{eq:xchange}
\delta(p_1,p_2) = n - \textnormal{CycleCount}(p_1,p_2) .
\end{equation*} 

\textbf{Cyclic edge distance and acyclic edge distance:}
\cite{ronald1997,ronald1995} defines the cyclic edge and acyclic edge distances 
for permutations that represent sets of edges. Element adjacency  
corresponds to undirected edges. Cyclic edge distance
considers the permutation to be a cycle, with adjacent endpoints; 
whereas acyclic edge distance does not.  
Cyclic edge distance interprets the permutation, $[0, 1, 2, 3]$,
as the undirected edges, $\{ (0, 1), (1, 2), (2, 3), (3, 0) \}$, while
acyclic edge distance excludes $(3, 0)$ from this set.
Both are invariant under a reversal (e.g., $[0, 1, 2, 3]$
is equivalent to $[3, 2, 1, 0]$). The cyclic form is also invariant under rotations. 
Distance is the number of edges in $p_1$ that are not in $p_2$, and is 
computed in $O(n)$ time.
Both are pseudo-metrics~\citep{ronald1997} due to reversal invariance, 
and rotational invariance for the cyclic form.
Define acyclic and cyclic edge distances, respectively, as:   
\begin{equation*}
\delta(p_1,p_2) = \lvert \{\begin{aligned}[t] & i \mid i\in\{1 \ldots n-1\} \, \wedge \\
	& \lvert p^{\unaryminus\!1}_2(p_1(i+1)) - p^{\unaryminus\!1}_2(p_1(i)) \rvert \neq 1 \} \rvert , \end{aligned}
\end{equation*}
\begin{equation*}
\delta(p_1,p_2) = \lvert \{\begin{aligned}[t] & i \mid i\in\{1 \ldots n\} \wedge x = \\
	& \lvert p^{\unaryminus\!1}_2(p_1((i\bmod n)+1)) - p^{\unaryminus\!1}_2(p_1(i)) \rvert \\
	& \wedge x \neq 1 \wedge x \neq n - 1 \} \rvert . \end{aligned} 
\end{equation*}

\textbf{R-type and cyclic r-type distances:}
The r-type distance (``r'' for relative)~\citep{campos2005}
is a directed edge version of acyclic edge distance.
Cyclic r-type~\citep{cicirello2016} is its cyclic counterpart,
including an edge between endpoints.
Though r-type is a metric, cyclic r-type 
is a pseudo-metric due to rotational invariance. Runtime for both is $O(n)$,
and defined respectively:
\begin{equation*}
\delta(p_1,p_2) = \lvert \{\begin{aligned}[t] & i \mid i\in\{1 \ldots n-1\} \, \wedge \\
	& p^{\unaryminus\!1}_2(p_1(i+1)) - p^{\unaryminus\!1}_2(p_1(i)) \neq 1 \} \rvert , \end{aligned}
\end{equation*}
\begin{equation*}
\delta(p_1,p_2) = \lvert \{\begin{aligned}[t] & i \mid i\in\{1 \ldots n\} \wedge x = \\
	& p^{\unaryminus\!1}_2(p_1((i\bmod n)+1)) - p^{\unaryminus\!1}_2(p_1(i)) \\
	& \wedge x \neq 1 \wedge x \neq 1 - n \} \rvert . \end{aligned}
\end{equation*}

\textbf{Kendall tau distance:}
The metric Kendall tau distance is a variation of Kendall's rank correlation 
coefficient~\citep{kendall1938}:
\begin{equation*}\label{eq:tau}
\delta(p_1,p_2) = \lvert \{\begin{aligned}[t] & x,y \in p_1 \mid p^{\unaryminus\!1}_1(x) > p^{\unaryminus\!1}_1(y) \\
	& \wedge p^{\unaryminus\!1}_2(x) < p^{\unaryminus\!1}_2(y) \} \rvert . \end{aligned}
\end{equation*}
Some divide by $n(n-1)/2$, but most use it as defined 
above~\citep{fagin2003,meila2010},
where it is the minimum number of adjacent swaps  
to transform $p_1$ into $p_2$, 
an adjacent swap edit distance.  
Compute in $O(n \lg n)$ time using a 
mergesort modified to count inversions.

\textbf{Reinsertion distance:}
An edit distance with a single 
atomic edit operation, removal/reinsertion, which
removes an element and reinserts it elsewhere in the permutation, is called
reinsertion distance. It is the minimum number of 
removal/reinsertions to transform $p_1$ into $p_2$.
Observing that the elements that must be removed and 
reinserted are exactly the elements that do not lie on the 
longest common subsequence~\citep{cicirello2016}, it is
computed efficiently in $O(n \lg n)$ time using \cite{hunt77}'s algorithm for
longest common subsequence.
We implement it as:
\begin{equation*}\label{eq:reinsert}
\delta(p_1,p_2) = n - \lvert\text{MaxCommonSubsequence}(p_1,p_2)\rvert .
\end{equation*}

\textbf{Deviation distance and normalized deviation distance:}
Deviation distance, a metric, sums the positional deviations
of the permutation elements. The positional deviation
of an element is the absolute difference of its index in $p_1$
from its index in $p_2$. \cite{ronald1998} originally divided this sum by $n-1$ 
to normalize an element's contribution to total distance in the interval $[0,1]$.
Many use this form~\citep{sorensen07,cicirello2013,cicirello2014}, while
others~\citep{sevaux2005,campos2005,cicirello2016} do not divide by $(n-1)$.
Runtime of our implementation is $O(n)$.
The two forms are:
\begin{equation*}
\delta(p_1,p_2) = \tfrac{1}{n-1} \sum_{e \in p_1} \lvert p^{\unaryminus\!1}_1(e)-p^{\unaryminus\!1}_2(e) \rvert , 
\end{equation*}
\begin{equation*}
\delta(p_1,p_2) = \sum_{e \in p_1} \lvert p^{\unaryminus\!1}_1(e)-p^{\unaryminus\!1}_2(e) \rvert .  
\end{equation*}

\textbf{Squared deviation distance:}
\cite{sevaux2005} suggest squared deviation distance, based on Spearman's rank 
correlation coefficient, 
summing of the squares of the positional deviations of the elements.
Prior authors state that squared deviation distance as well as deviation 
distance require quadratic time~\citep{sevaux2005}.
However, we implement these in $O(n)$ time with two linear passes, 
one computing the inverse of one permutation,
which is then used in the second pass to look up element indexes.
\begin{equation*}
\delta(p_1,p_2) = \sum_{e \in p_1} (p^{\unaryminus\!1}_1(e)-p^{\unaryminus\!1}_2(e))^2 . 
\end{equation*}

\textbf{Lee distance:}
Lee distance originated in coding theory for strings~\citep{lee58}. Here we adapt it
for permutation distance.
Lee distance, a metric, treats the permutation as a cycle, summing
the minimum of the left and right positional deviations.
It is computed in $O(n)$ time.
Define as:
\begin{equation*}
\delta(p_1,p_2) = \sum_{e \in p_1} \min( \begin{aligned}[t] & \lvert p^{\unaryminus\!1}_1(e)-p^{\unaryminus\!1}_2(e) \rvert, \\ 
	& n - \lvert p^{\unaryminus\!1}_1(e)-p^{\unaryminus\!1}_2(e) \rvert ) . \end{aligned}
\end{equation*}

\textbf{Reversal edit distance:}
Reversal edit distance is the minimum number of reversals to transform $p_1$ into $p_2$.
Computing reversal edit distance is NP-Hard~\citep{caprara1997}; and  
\cite{schiavinotto2007} argue that the best available approximations are 
insufficient for search landscape analysis.
We implement it with a breadth-first enumeration to
initialize a lookup table mapping each of the $n!$ permutations to its reversal 
edit distance from a reference permutation.
Computing the distance between a given pair of permutations is then a table lookup.
This is only feasible for short permutation length.
Initialization cost is $O(n! n^3)$ (i.e., breadth-first enumeration of $O(n!)$ permutations, 
each with $O(n^2)$ neighbors (possible sub-permutation reversals), and a reversal costs $O(n)$).  
Applications with the need to compute $O(n!)$ distances all from the same reference permutation 
can do so with an amortized initialization cost of $O(n^3)$ per distance calculation. 
The table lookup costs $O(n^2)$ (cost to compute mixed radix representation of the permutation).

\section{Classifying Distance Metrics}\label{sec:classes}

We perform PCA to identify groups of related permutation metrics, using 
the distance measures from Section~\ref{sec:distance} except 
the edit, normalized deviation, and reversal edit distances.  
Normalized deviation distance is deviation distance scaled,
and thus observations made about one apply to both.
Edit distance's parameters define a continuum
of metrics. Reversal edit distance is too costly, 
but we later discuss how it fits in our classification.

\begin{table*}[t]
\begin{center}
\begin{minipage}{\textwidth}
\caption{Lower triangle of correlation matrix (columns in same order as rows)}\label{tab:corr}
\begin{tabular*}{\textwidth}{@{\extracolsep{\fill}}lrrrrrrrrrrr@{\extracolsep{\fill}}}
\toprule
exact match	& 1.0&&&&&&&&&&\\
interchange	& .766	& 1.0&&&&&&&&&\\
acyclic edge	& .019	& .070	 &1.0&&&&&&&&\\
cyclic edge	&-.000	& .056	& .899	& 1.0&&&&&&&\\
rtype	 	& .024	& .009	& .628	& .564	& 1.0&&&&&&\\
cyclic rtype	& -.000&	-.010	& .557	& .619	& .911	& 1.0&&&&&\\
Kendall tau	& .328	& .241	&-.000	& .000	& .085	& .075	& 1.0&&&&\\
reinsertion	& .301	& .182	 &.102	& .100	& .422	& .392	& .704&	 1.0&&&\\
deviation	& .515	 &.395	 &.008	&-.000	 &.020	&-.000	& .931	& .650	& 1.0&&\\
sq. deviation	& .333	& .255	&-.000&	-.000	& .017	&-.000	 &.984	 &.623	 &.947	 &1.0&\\
Lee	 		& .556	& .426	& .019	& .000	& .014	& -.000 & .447 & .452 & .703 & .455 & 1.0 \\
\botrule
\end{tabular*}
\end{minipage}
\end{center}
\end{table*}

For all permutations of length $n=10$, compute 
distance to a single reference permutation. Using Jacobi iteration, 
compute the eigenvalues and eigenvectors of the correlation matrix of 
Table~\ref{tab:corr}. Table~\ref{tab:eigenvalues} lists 
the eigenvalues of the principal components (PC), the first three of which 
are greater than 1.0; and the first five PCs combine for more than 90\% of the sum.  
Table~\ref{tab:eigenvectors} shows the eigenvectors for the first five PCs.
Correlation between the original distance metrics 
and each of the first five PCs is in Table~\ref{tab:dist-pc-corr}. The 
first three PCs correspond to the three types of permutation problem 
discussed earlier in Section~\ref{sec:intro}.

\begin{table}[t]
\begin{center}
\begin{minipage}{216pt} %{174pt} 
\caption{Eigenvalues of the PCs}\label{tab:eigenvalues}
\begin{tabular}{@{}rrrr@{}}
\toprule
PC &	eigenvalue	& proportion	& cumulative \\ 
\midrule
  1	& 4.3644	& 	0.3968	& 	0.3968\\
  2	& 	3.1148	& 	0.2832	& 	0.6799\\
  3	& 	1.4740	& 	0.1340	& 	0.8139\\
  4	& 	0.8367	& 	0.0761	& 	0.8900\\
  5	& 	0.5465	& 	0.0497	& 	0.9397\\
  6	& 	0.2492	& 	0.0227	& 	0.9623\\
  7	& 	0.2120	& 	0.0193	& 	0.9816\\
  8	& 	0.1575	& 	0.0143		& 0.9959\\
  9	& 	0.0315	& 	0.0029		& 0.9988\\
 10	& 	0.0107		& 0.0010		& 0.9998\\
 11	& 	0.0026		& 0.0002		& 1.0000\\
\botrule
\end{tabular}
\end{minipage}
\end{center}
\end{table}

\begin{table}[t]
\begin{center}
\begin{minipage}{216pt} %{174pt}
\caption{Eigenvectors of the first five PCs}\label{tab:eigenvectors}
\begin{tabular}{@{}lrrrrr@{}}
\toprule
distance&	PC1	&PC2	&PC3	&PC4	&PC5 \\
\midrule
exact match		&	.2984&	.0958	&.5419	&-.1573	&.1423	\\
interchange		&	.2487&	.0695	&.6058	&-.0586	&.3936	\\
acyclic edge		&	.0854&	-.4751	&.1354	&.4611	&-.0635	\\
cyclic edge	&	.0805&	-.4768	&.1194	&.4674	&-.0455	\\
r-type	&	.1271&	-.4873	&-.0576&	-.3803	&.0517	\\
cyclic r-type		&	.1153&	-.4874	&-.0666&	-.3793	&.0510	\\
Kendall tau	&	.4216&	.0928	&-.3110&	.1400	&.2292	\\
reinsertion		&	.3721&	-.0848	&-.2529&	-.3795	&-.0509	\\
deviation	&	.4516&	.1321	&-.1089&	.1630	&-.0651	\\
sq. deviation	&	.4140&	.1189	&-.2828&	.2444	&.2218	\\
Lee			&	.3381&	.1027	&.2157	&-.0476	&-.8396	\\
\botrule
\end{tabular}
\end{minipage}
\end{center}
\end{table}

\begin{table}[t]
\begin{center}
\begin{minipage}{216pt} %{174pt}
\caption{Correlation between distances and first five PCs}\label{tab:dist-pc-corr}
\begin{tabular}{@{}lrrrrr@{}}
\toprule
distance&	PC1	&PC2	&PC3	&PC4	&PC5 \\
\midrule
exact match		&	.6234	& .1691	& .6579	& -.1439	& .1052	\\
interchange		&	.5196	& .1227	& .7355	& -.0536	& .2910	\\
acyclic edge		&	.1784	& -.8385	& .1644	& .4218	& -.0470		\\
cyclic edge	&	.1682	& -.8415	& .1450	& .4276	& -.0337	\\
r-type		&	.2654	& -.8600	& -.0699	& -.3479	& .0382	\\
cyclic r-type		&	.2410	& -.8602	& -.0808	& -.3469	& .0377	\\
Kendall tau	&	.8808	& .1638	& -.3775	& .1281	& .1695	\\
reinsertion		&	.7774	& -.1497	& -.3070	& -.3472	& -.0377	\\
deviation	&	.9435	& .2332	& -.1322	& .1491	& -.0481	\\
sq. deviation	&	.8649	& .2099	& -.3434	& .2236	& .1640	\\
Lee			&	.7063	& .1812	& .2619	& -.0436	& -.6207	\\
\botrule
\end{tabular}
\end{minipage}
\end{center}
\end{table}

\textbf{PC1 (P-permutation):} 
PC1 correlates extremely strongly (0.94) to deviation distance, very strongly
to the Kendall tau and squared deviation distances, and reasonably strongly 
to the reinsertion and Lee distances (Table~\ref{tab:dist-pc-corr}).  
The Kendall tau and reinsertion distances, by their definitions, measure similarity
in terms of pairwise element precedences.  
The variations of deviation distance capture that same essence in that 
an element that is displaced a greater number of positions
is likely involved in a greater number of precedence 
inversions (i.e., where $a$ is prior to $b$
in one permutation, and somewhere after in the other).  
These five permutation metrics are P-permutation distances,
measuring permutation distance in terms of precedence related features.

\textbf{PC2 (R-permutation):} 
PC2 correlates very strongly with both forms of edge distance, 
and both forms of R-type distance ($\lvert r \rvert > 0.83$ in all four cases).  
These distances all focus on adjacency (i.e., edges) of permutation elements.

\textbf{PC3 (A-permutation):} PC3 strongly correlates to the exact match ($r=0.6579$) and interchange distances  
($r=0.7355$). Both focus on absolute positions of permutation elements.

The fourth and fifth PCs identify subtypes. Their eigenvalues are less 
than 1, and account for small portions of the eigenvalue 
sum (7.6\% and 5\%), but interpreting is interesting none-the-less.

\textbf{PC4 (R-permutation, undirected):} PC4's 
strongest correlations are to the two variations of edge distance,
which consider permutations to represent sets of undirected edges.

\textbf{PC5 (P-permutation, cyclic):} 
PC5 correlates strongly ($\lvert r \rvert = 0.6207$) to Lee distance, 
and only weakly to the others.  
Lee distance also strongly correlates with PC1 (P-permutation), but 
is different than the other deviation-based metrics 
in that it computes positional deviation as if the end points are linked.  
In some sense, this is a cyclic subtype of P-permutation.

Table~\ref{tab:categories} summarizes this classification.
There are three primary types: P-permutation, R-permutation, and A-permutation; and
two are decomposed into subtypes. Although we excluded reversal edit distance in the analysis,
we include it among the undirected R-permutation metrics as a reversal clearly replaces two
undirected edges.

\begin{table}[t]
\begin{center}
\begin{minipage}{216pt} %{174pt}
\caption{Permutation distance metric classification}\label{tab:categories}
\begin{tabular}{@{}lll@{}}
\toprule
type			& subtype				& distances	\\ 
\midrule
\multirow{4}{*}{P-permutation}	& acyclic		& Kendall tau, reinsertion, \\
				&						& deviation, \\
				&						& squared deviation \\\cdashline{2-3}
				& cyclic		& Lee \\ 
\midrule
\multirow{3}{*}{R-permutation}	& undirected	& acyclic edge, cyclic edge, \\
				&						& reversal edit \\\cdashline{2-3}
				& directed		& r-type, cyclic r-type \\ 
\midrule
A-permutation	&						& exact match, interchange \\ 
\botrule
\end{tabular}
\end{minipage}
\end{center}
\end{table}

\section{On Effects of Length}\label{sec:longperms}

To derive the classification in Section~\ref{sec:classes}, 
the PCA computed the correlation matrix via brute-force enumeration of
permutations of length $n=10$. To explore whether
length affects the classes identified, we repeat the PCA using length $n=50$,
which is too long for exhaustive enumeration. We instead randomly sample the space of
permutations, sampling 3628800 random permutations of length 50, chosen so number of data 
points is the same as Section~\ref{sec:classes}. Table~\ref{tab:corr50} shows the
correlation matrix. Table~\ref{tab:eigenvalues50} provides the eigenvalues, 
and Table~\ref{tab:eigenvectors50} shows the eigenvectors of the first five PCs.
Table~\ref{tab:dist-pc-corr50} shows the correlations between the original 
metrics and the first five PCs.

\begin{table*}[t]
\begin{center}
\begin{minipage}{\textwidth}
\caption{Lower triangle of correlation matrix (permutation length 50)}\label{tab:corr50}
\begin{tabular*}{\textwidth}{@{\extracolsep{\fill}}lrrrrrrrrrrr@{\extracolsep{\fill}}}
\toprule
exact match	& 1.0 &&&&&&&&&&\\
interchange	& .578	& 1.0 &&&&&&&&&\\
acyclic edge	& .001	& .009	& 1.0 &&&&&&&&\\
cyclic edge	& .000	& .009	& .980	& 1.0 &&&&&&&\\
rtype	& .001	& .000	& .693	& .679	& 1.0 &&&&&&\\
cyclic rtype	&-.000	&-.000	& .679	& .693	& .980	& 1.0 &&&&&\\
Kendall tau	& .142	& .082	&-.000	&-.000	& .008	& .007	& 1.0 &&&&\\
reinsertion	& .140	& .074	& .060	& .059	& .176	& .172	& .532	& 1.0 &&&\\
deviation	& .226	& .132	&-.000	&-.000	&-.001	&-.001	& .944	& .555	& 1.0 &&\\
sq. deviation	& .143	& .084	&-.000	&-.000	&-.000	&-.001	& .995	& .501	& .949	& 1.0 &\\
Lee	& .248	& .144	& .000	&-.000	&-.001	&-.001	& .431	& .439	& .685	& .433	& 1.0 \\
\botrule
\end{tabular*}
\end{minipage}
\end{center}
\end{table*}

\begin{table}[t]
\begin{center}
\begin{minipage}{216pt} %{174pt}
\caption{Eigenvalues of the PCs (permutation length 50)}\label{tab:eigenvalues50}
\begin{tabular}{@{}rrrr@{}}
\toprule
PC &	eigenvalue	& proportion	& cumulative \\ 
\midrule
  1	&3.7755	&0.3432	&0.3432\\
  2	&3.3513	&0.3047	&0.6479\\
  3	&1.5162	&0.1378	&0.7857\\
  4	&0.7515	&0.0683	&0.8541\\
  5	&0.6604	&0.0600	&0.9141\\
  6	&0.4849	&0.0441	&0.9582\\
  7	&0.4111	&0.0374	&0.9955\\
  8	&0.0336	&0.0031	&0.9986\\
  9	&0.0069	&0.0006	&0.9992\\
 10	&0.0059	&0.0005	&0.9998\\
 11	&0.0027	&0.0002	&1.0000\\
\botrule
\end{tabular}
\end{minipage}
\end{center}
\end{table}

\begin{table}[t]
\begin{center}
\begin{minipage}{216pt} %{174pt}
\caption{Eigenvectors of first five PCs (length 50)}\label{tab:eigenvectors50}
\begin{tabular}{@{}lrrrrr@{}}
\toprule
distance&	PC1	&PC2	&PC3	&PC4	&PC5 \\
\midrule
exact match		&	-.160	&-.039	&-.671	&.011	&.019\\
interchange		&	-.113	&-.025	&-.692	&.151	&.159\\
acyclic edge		&	-.095	&.488	&-.011	&.242	&-.335\\
cyclic edge	&	-.095	&.488	&-.010	&.242	&-.335\\
r-type	&	-.109	&.488	&.006	&-.194	&.309\\
cyclic r-type		&	-.108	&.488	&.006	&-.192	&.308\\
Kendall tau	&	-.468	&-.110	&.163	&.309	&.166\\
reinsertion		&	-.355	&.002	&.063	&-.556	&.290\\
deviation	&	-.492	&-.119	&.083	&.099	&-.099\\
sq. deviation	&	-.465	&-.113	&.161	&.336	&.142\\
Lee			&-.343	&-.081	&-.082	&-.509	&-.647	  \\
\botrule
\end{tabular}
\end{minipage}
\end{center}
\end{table}

\begin{table}[t]
\begin{center}
\begin{minipage}{216pt} %{174pt}
\caption{Correlation between distance and first five PCs.}\label{tab:dist-pc-corr50}
\begin{tabular}{@{}lrrrrr@{}}
\toprule
distance&	PC1	&PC2	&PC3	&PC4	&PC5 \\
\midrule
exact match	&-.311	&-.072	&-.827	&.009	&.016		\\
interchange	&-.219	&-.046	&-.853	&.131	&.129		\\
acyclic edge	&-.185	&.893	&-.013	&.209	&-.272			\\
cyclic edge	&-.185	&.893	&-.013	&.210	&-.272	\\
r-type			&-.212	&.893	&.007	&-.169	&.251\\
cyclic r-type	&-.211	&.893	&.007	&-.167	&.250		\\
Kendall tau	&-.908	&-.202	&.201	&.268	&.135	\\
reinsertion	&-.690	&.003	&.078	&-.482	&.236		\\
deviation		&-.956	&-.218	&.102	&.086	&-.080\\
sq. deviation	&-.903	&-.207	&.198	&.291	&.115	\\
Lee	&-.666	&-.149	&-.100	&-.441	&-.526			\\
\botrule
\end{tabular}
\end{minipage}
\end{center}
\end{table}

In Table~\ref{tab:dist-pc-corr50}, PC1 again corresponds to the P-permutation metrics,
correlating extremely strongly to the deviation, Kendall tau, and squared deviation 
distances ($\lvert r \rvert>0.9$), and also correlating strongly to the reinsertion 
and Lee distances. Likewise, PC2 (as before) corresponds to the
R-permutation metrics, with very strong correlation ($\lvert r \rvert>0.89$)
to both forms of edge distance and both forms of R-type distance. Also consistent
with the shorter length results, PC3 corresponds to the A-permutation metrics,
with very strong correlation to the exact match ($\lvert r \rvert = 0.8265$) and
interchange distances ($\lvert r \rvert = 0.8525$). PC5 again correlates moderately 
to Lee distance ($\lvert r \rvert = 0.5258$) and only weakly to the others.

PC4 is the only inconsistency between the longer permutation results 
and the shorter permutation results. With shorter permutations, PC4 
identified the two forms of edge distance (R-permutation undirected subtype).  
With longer randomly sampled permutations, PC4 identified reinsertion distance, and to a lesser extent
Lee distance. This suggests that as length increases there may be a relationship
between the reinsertion and Lee distances; or that reinsertion distance captures
a different form of variability than does the other P-permutation metrics.

We stick with the earlier classification (Table~\ref{tab:categories}), since
four PCs directly correspond to that analysis, and since the distinct
nature of reinsertion distance is not entirely clear.

\section{Example Fitness Landscapes}\label{sec:landscapes}

We present five search landscapes as examples.

\textbf{R-permutation (undirected) landscape ($L_1$):} The first search landscape
is a simple instance of the TSP with a known optimal solution. It has 20 cities 
arranged equidistantly on a circle of radius 1.0. The edge cost is Euclidean 
distance. The optimal solution is to either follow the cities clockwise or 
counterclockwise around the circle. There are 40 optimal permutations: 20 cities 
at which to begin, and two possible travel directions (clockwise and counterclockwise).

We compute FDC using 100000 randomly sampled
permutations (Table~\ref{tab:landscapes}). In this case, it is the correlation
between tour cost, and distance to the nearest of the 40 optimal permutations. 
The highest FDC is for the two forms of edge distance, followed by
the two forms of R-type (recall that R-type distance uses directed edges, while
edge distance uses undirected edges). Cyclic edge distance has slightly higher FDC
over acyclic edge distance, which makes sense since a solution to a TSP is a cycle of the
cities where the first and last permutation elements represent an edge.

\begin{table}[t]
\begin{center}
\begin{minipage}{216pt} %{174pt}
\caption{FDC for the metrics and example landscapes}\label{tab:landscapes}
\centering
\begin{tabular}{@{}lrrrrr@{}}
\toprule
distance	& $L_1$	& $L_2$	& $L_3$	& $L_4$	& $L_5$ \\ 
\midrule
exact match		&.1548		&.1881		& .6917	&.2974	&.4806\\
interchange		&.1192		&.0886		& .5296	&.2204	&.3665\\
acyclic edge		&.6052	&.3474		&.0118		&.0020	&.0186\\
cyclic edge		&.6204	&.3822		&-.0002	&.0006	&.0026\\
r-type				&.5442 & .6333	&.0148		&.0790	&.0136\\
cyclic r-type		&.5562 & .6595	&-.0016	&.0684	&.0005\\
Kendall tau		&.3423		&.2408		&.2245		& .9022	&.3862\\
reinsertion		&.3382		&.5349		&.2080		& .6364	&.3887\\
deviation			&.3898		&.1875		&.3544		& .8410	&.6072\\
sq. deviation	&.3150		&.1555		&.2282		& .8876	&.3935\\
Lee				&.4640		&.2316		&.3836		&.4063	& .8619 \\
\botrule
\end{tabular}
\end{minipage}
\end{center}
\end{table}

\textbf{R-permutation (directed) landscape ($L_2$):}
This landscape is a simple Asymmetric TSP (ATSP), with
20 cities arranged equidistantly on a circle of radius 1.0. Let 
city $c_i$ be the city after $c_{i-1}$ in counterclockwise order.
The edge cost from $c_i$ to $c_j$ is Euclidean distance if $i < j$,
and is otherwise a constant distance 2.0. There is one optimal
tour, which visits the cities in counterclockwise order, and
20 permutations that correspond to this tour (20 starting cities).

For this landscape, the two forms of R-type distance offer the highest FDC
(Table~\ref{tab:landscapes}), and FDC is 
low for the other distance measures. Consistent with the cyclic nature of 
the ATSP, cyclic R-type has slightly higher FDC than the acyclic form. 

\textbf{A-permutation landscape ($L_3$):}
Landscape $L_3$ is a variation of the {\em Permutation in a Haystack} 
problem~\citep{cicirello2016}, where
one must minimize distance to a predetermined optimal permutation $p$. 
It is the permutation analog of the OneMax~\citep{Ackley1985} 
bitstring problem often used in testing genetic algorithms. The Permutation 
in a Haystack enables defining
fitness landscapes with a known optimal solution that possess a desired 
topology, via choice of distance function.

For $L_3$, we modify the Permutation in a Haystack to 
use a noisy distance function as the optimization objective. Given target $p$, 
the fitness of permutation $q$ is $\alpha_q * \delta(p, q)$, where
$\delta(p, q)$ is exact match distance between $q$ and optimal solution $p$,
and each $\alpha_q$ is uniformly random from the interval $[1, 1.5)$. 
We use a smaller permutation length of 10 for $L_3$ than we did for the first two. 

In Table~\ref{tab:landscapes}, we see that the two metrics, earlier
identified as A-permutation by our PCA, both have high FDC to landscape $L_3$; and FDC
is low for all other permutation metrics.

\textbf{P-permutation (acyclic) landscape ($L_4$):}
We modify the noisy Permutation in a Haystack from $L_3$
to obtain a P-permutation landscape, $L_4$, with a known optimal solution
by using the Kendall tau distance instead of exact match.

Three of the four metrics classified as acyclic 
P-permutation have very high FDC for 
landscape $L_4$ (Table~\ref{tab:landscapes}), including the Kendall tau, 
deviation, and squared deviation distances. The fourth, reinsertion 
distance, also has reasonably high FDC for $L_4$;
while FDC is low for all other metrics.

\textbf{P-permutation (cyclic) landscape ($L_5$):}
The last example fitness landscape uses Lee distance in the
noisy Permutation in a Haystack adapted from
$L_3$ and $L_4$. Lee distance provides the 
highest FDC for this landscape (Table~\ref{tab:landscapes}).

\section{Mutation Classification}\label{sec:mutation}

The metrics with highest FDC for each fitness landscape (Section~\ref{sec:landscapes}) correspond 
to the metrics for the landscape's class in Table~\ref{tab:categories}. We now explore 
how a corresponding mapping of mutation operators for an EA can assist in selecting relevant 
operators for the optimization problem at hand.

We explore the behavior of a variety of mutation operators for
problems of each type. All test problems
are over the space of permutations of length 100. The EA is a mutation-only generational
model such that children replace the parents in the population. We use elitism, keeping the
best population member unaltered, and otherwise use Stochastic Universal 
Sampling (SUS)~\citep{Baker1987} for selection. 
All problems involve minimizing the cost $c(p)$ of permutation $p$. The EA defines fitness of $p$
as: $f(p) = 1 / (1 + c(p))$. The permutations selected by SUS in each generation undergo
one mutation each. Population size is 100. 
Results are averages of 100 runs on 100 instances.

We compare several mutation operators, summarized with runtimes in Table~\ref{tab:mutation}.
Adjacent Swap (AdjSwap) swaps 
a random pair of adjacent elements. Swap exchanges the positions of two random
elements. Insertion removes a random element and reinserts it at a random index.
Reversal reverses a random segment. 3opt assumes the permutation 
represents a set of edges and replaces three random edges~\citep{Lin1965}.
BlockMove removes a random segment and reinserts it at a random location.
BlockSwap swaps the locations of two random segments. $\mathrm{Cycle}(\mathit{kmax})$ 
mutation, one of two forms of cycle mutation~\citep{cicirello2022}, induces a random $k$-cycle 
for random $k \in [2,\mathit{kmax}]$, where we use $\mathit{kmax}=10$
in the experiments. Scramble randomizes the order of a random contiguous segment. 

\begin{table}[t]
\begin{center}
\begin{minipage}{216pt} %%{174pt}
\caption{Summary of mutation operators}\label{tab:mutation}
\begin{tabular}{@{}ll@{}}
\toprule
Mutation Operator		& Runtime \\ 
\midrule
AdjSwap	& $O(1)$ \\
Swap	& $O(1)$ \\
Insertion	& $O(n)$ \\
Reversal	& $O(n)$ \\
3opt	& $O(n)$ \\
BlockMove	& $O(n)$ \\
BlockSwap	& $O(n)$ \\
Cycle	& $O(\min(n,k^2))$ \\
Scramble	& $O(n)$ \\
Uniform	& $O(n)$ \\ 
\botrule
\end{tabular}
\end{minipage}
\end{center}
\end{table}

Uniform Scramble, believed to be new with this paper, randomizes the positions
of a random set of elements, each element chosen with probability $U$.
In the experiments, $U=1/3$ to affect the same number of elements
on average as Scramble.

\begin{figure}[t]
\centering
\includegraphics{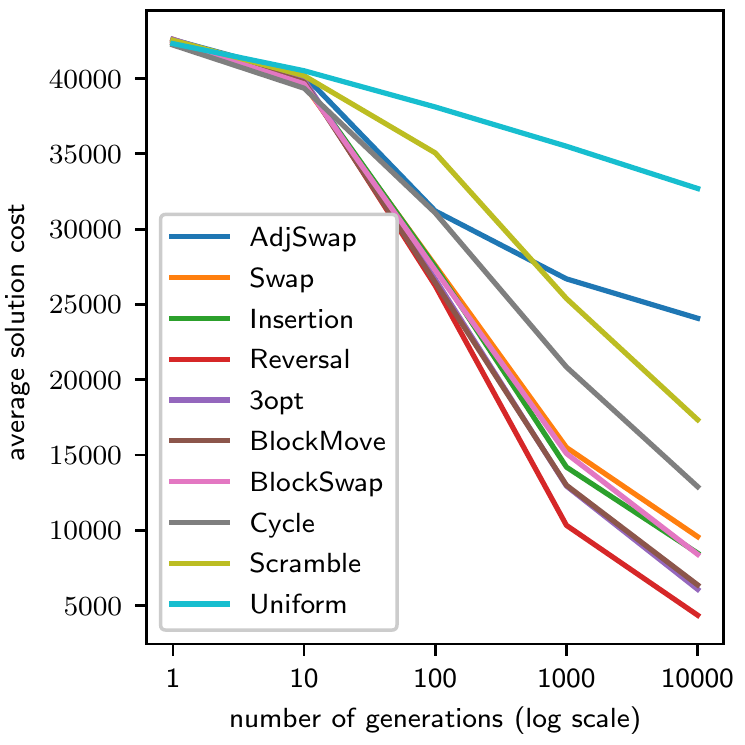}
\caption{Mutation operator comparison for the TSP}\label{fig:tsp}
\end{figure}

\textbf{R-permutation (undirected) problem:} In Figure~\ref{fig:tsp}, we see results
for TSP instances generated from random symmetric distance
matrices. The $x$-axis is number of generations at log scale. Reversal, 
equivalent to randomly replacing two undirected edges, performs best.
Next best are BlockMove and 3opt (both replace three edges), followed closely by 
Insertion (changes three edges), BlockSwap and Swap (both replace four edges). 
The others lag behind, although Cycle is tunable via the maximum cycle length.

\begin{figure}[t]
\centering
\includegraphics{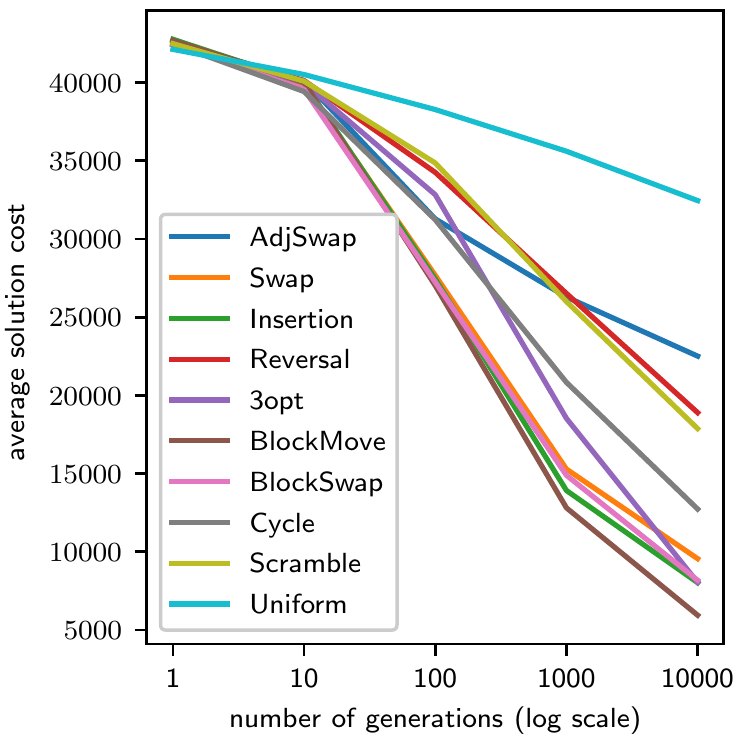}
\caption{Mutation operator comparison for the ATSP}\label{fig:atsp}
\end{figure}

\textbf{R-permutation (directed) problem:} 
For ATSP instances generated from random asymmetric distance
matrices (Figure~\ref{fig:atsp}), the best performing mutation is BlockMove, followed by Insertion,
BlockSwap, and Swap. At 10000 generations, 3opt achieves
equivalent performance, but under-performs earlier. The best performing
mutations are similar to the undirected case, except
Reversal performs poorly for directed edges, likely because
reversing a segment flips the direction of $3n$ directed 
edges on average. This explains 3opt's mixed 
performance (some 3opt moves flip direction of many edges,
while others do not).

\begin{figure}[t]
\centering
\includegraphics{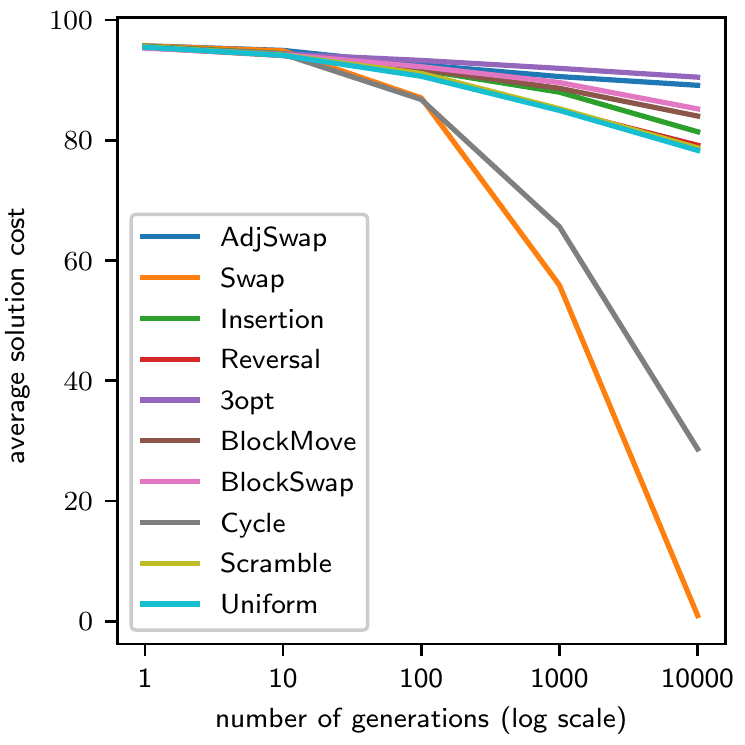}
\caption{Mutation operator comparison for the Permutation in a Haystack with Exact Match distance}\label{fig:em}
\end{figure}

\textbf{A-permutation problem:} The Permutation in a Haystack
with Exact Match distance provides an A-permutation landscape (the 
original version, and not the noisy version 
of Section~\ref{sec:landscapes}). Swap clearly dominates (Figure~\ref{fig:em}).
The only other mutation worth considering for A-permutation problems is Cycle 
mutation, whose cycle length parameter may need fine-tuning.

\begin{figure}[t]
\centering
\includegraphics{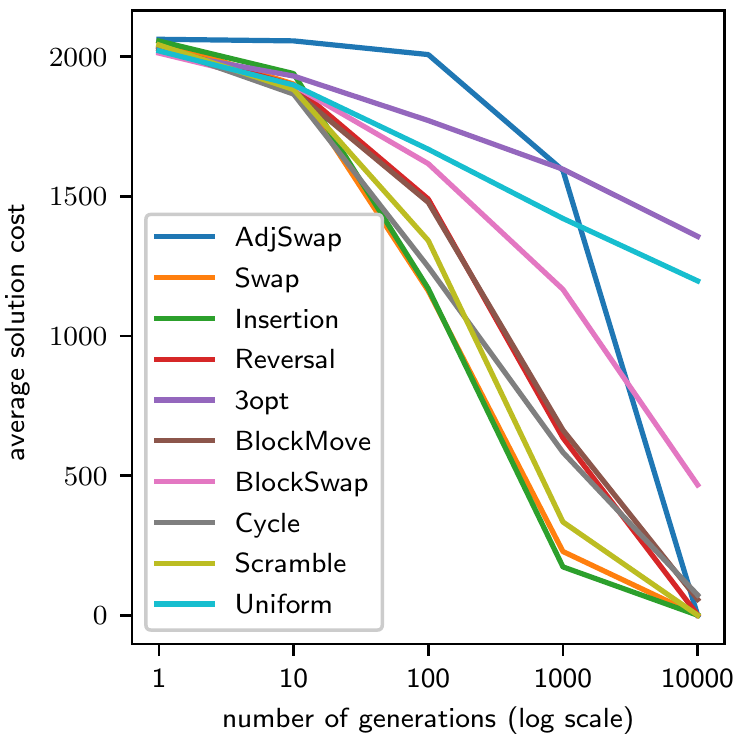}
\caption{Mutation operator comparison for the Permutation in a Haystack with Kendall tau distance}\label{fig:tau}
\end{figure}

\textbf{P-permutation (acyclic) problem:} We use the Permutation in a Haystack with Kendall tau
distance to create an acyclic P-permutation problem (results in Figure~\ref{fig:tau}). Many of the mutation
operators are able to consistently optimally solve this problem for 10000 generation runs. Thus, we focus
on earlier performance. At 1000 generations and earlier, the best performing mutation is Insertion,
followed closely by Swap, and then most surprisingly by Scramble. The others lag behind.

\begin{figure}[t]
\centering
\includegraphics{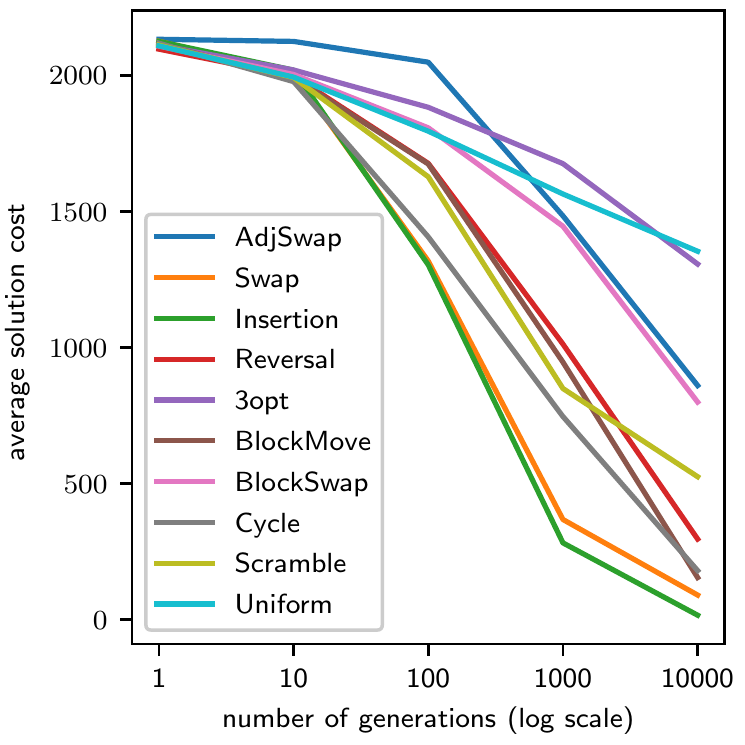}
\caption{Mutation operator comparison for the Permutation in a Haystack with Lee distance}\label{fig:lee}
\end{figure}

\textbf{P-permutation (cyclic) problem:} The Permutation in a Haystack with Lee distance
creates a cyclic P-permutation problem. Insertion and Swap are
the only operators that consistently perform well throughout the run in this case (Figure~\ref{fig:lee}).

From the computational results, we derive a classification of the permutation mutation
operators (Table~\ref{tab:mutationclasses}) as a counterpart to the 
classification of permutation metrics. The two schemes can be used in
combination to inform the design of search algorithms. When designing an EA, or other metaheuristic, 
an analysis of the fitness landscape for the problem can utilize the classes of
distance functions (Table~\ref{tab:categories}) to identify the permutation features with greatest 
impact on fitness. The mutation operator classification (Table~\ref{tab:mutationclasses}) then 
provides a catalog of mutation operators that are well-suited to optimizing the problem at hand.

\begin{table}[t]
\begin{center}
\begin{minipage}{216pt} %{174pt}
\caption{Mutation operator classification}\label{tab:mutationclasses}
\begin{tabular}{@{}lll@{}}
\toprule
type			& subtype				& mutation operators	\\ 
\midrule
\multirow{2}{*}{P-permutation}	& acyclic		& Insertion, Swap, Scramble \\ \cdashline{2-3}
				& cyclic		&  Insertion, Swap \\ 
\midrule
\multirow{5}{*}{R-permutation}	& undirected	& Reversal, BlockMove,   \\
				&					& 3opt, Insertion,  \\
				&					& BlockSwap, Swap \\\cdashline{2-3}
				& directed		& BlockMove, Insertion, \\
				&				& BlockSwap, Swap \\ 
\midrule
A-permutation	&						& Swap, Cycle \\ 
\botrule
\end{tabular}
\end{minipage}
\end{center}
\end{table}

\section{Discussion and Conclusions}\label{sec:conclude}

Our first contribution is the classification
of permutation metrics according to the features that
influence each measure of distance. We derived the classification formally
using PCA, and included all of the common measures of permutation distance,
as well as a few less common metrics.

The classification aligns with an existing set of less formally
derived problem classes, which leads to our next contribution, formal confirmation
of those problem classes, in that FDC is highest for the metrics that correspond
to a problem's class. Our PCA approach also lead to identifying new problem subtypes.
With this alignment between metric classes and problem classes, the results aid in
selecting problem-relevant metrics for use with fitness landscape analysis techniques, 
such as FDC. For example, if faced with a problem where 
permutations represent sets of edges (e.g., TSP), then the classification suggests using 
one of the forms of edge or r-type distance, depending upon whether edges are 
undirected or directed. If it is a P-permutation problem, where general pairwise 
element precedences have greatest impact on fitness (e.g., many scheduling problems), 
then you would choose a P-permutation metric such as Kendall tau, reinsertion, 
or one of the variations of deviation distance. Additionally, in this case, 
you may then factor in the runtimes of the metrics. For example, the Kendall tau and squared 
deviation distances correlate very strongly ($r=0.984$, Table~\ref{tab:corr}). However, 
Kendall tau is computed in $O(n \lg n)$ time, while squared deviation is computed 
in $O(n)$ time. Even if Kendall tau is the better match for your specific problem, 
squared deviation may be sufficient due to its strong correlation while saving computational 
cost.

Our third contribution is a comprehensive classification of permutation mutation operators, 
as a counterpart to the distance classification, that can inform the effective selection of 
a mutation operator during the design phase of an EA. The distance classes 
support fitness landscape analysis and problem class identification. The mutation operator 
classification then narrows in on the most relevant available mutation operators for the given 
problem class. We included all of the most common, as well as less common, mutation operators 
for permutations in this research. 

This paper also contributes a new mutation operator,
Uniform Scramble, which is a variation of 
Scramble. Although Uniform Scramble was not among the top-performing 
for any problem class, it has a tunable parameter that will be explored in 
future work, or it may be useful to give a stagnated search a
kick (e.g., it is disruptive).

All of the distance metrics are implemented in the open-source JavaPermutationTools (JPT)
library~\citep{cicirello2018}. JPT's source code is available 
at: \url{https://github.com/cicirello/JavaPermutationTools}. The EA implementation and
all of the mutation operators are in the open-source Chips-n-Salsa library~\citep{cicirello2020},
whose source code is available at: \url{https://github.com/cicirello/Chips-n-Salsa}. Source
code to replicate the analysis, experiments, and results of 
Sections~\ref{sec:classes}--\ref{sec:mutation} is available
at: \url{https://github.com/cicirello/MONE2022-experiments}.

\backmatter

\section*{Declarations}

\noindent \textbf{Funding:} Not applicable.

\bigskip

\noindent \textbf{Conflict of Interest / Competing Interests:} The author declares that he has no conflict of interest.

\bigskip

\noindent \textbf{Ethics Approval:} Not applicable.

\bigskip

\noindent \textbf{Availability of Data:} Raw and processed data produced by the experiments is 
available: \url{https://github.com/cicirello/MONE2022-experiments}.

\bigskip

\noindent \textbf{Code Availability:} Code available at: \url{https://github.com/cicirello/MONE2022-experiments}.

\bibliography{mone2022}% common bib file

\end{document}